\pdfoutput=1

\documentclass{article}
\usepackage{ijcai21}

\usepackage{times}
\usepackage{soul}
\usepackage{url}
\usepackage[hidelinks]{hyperref}
\usepackage[utf8]{inputenc}
\usepackage[small]{caption}
\usepackage{graphicx}
\usepackage{amsmath}
\usepackage{amsthm}
\usepackage{booktabs}
\usepackage{algorithm}
\usepackage{stackengine}
\usepackage{algorithmic}
\usepackage{xcolor}
\usepackage[export]{adjustbox}
\usepackage{multicol}
\usepackage{multirow}
\usepackage{makecell}

\usepackage{tikz}
\newcommand*\circled[1]{\tikz[baseline=(char.base)]{
            \node[shape=circle,draw,inner sep=0,minimum size=0.3cm] (char) {\scriptsize{#1}};}}

\DeclareMathOperator{\Score}{Score}
\DeclareMathOperator{\Normalize}{Norm}

\title{Federated Face Recognition\footnote{This paper was submitted to IJCAI 2021 on January 20, 2021, but was unfortunately rejected. }}

\author{
Fan Bai$^1$\and
Jiaxiang Wu$^2$\and
Pengcheng Shen$^2$\and
Shaoxin Li$^2$\And
Shuigeng Zhou$^1$
\affiliations
$^1$Fudan University\\
$^2$Youtu Lab, Tencent
\emails
\{fbai19, sgzhou\}@fudan.edu.cn,
\{willjxwu, quantshen, darwinli\}@tencent.com
}

\begin{document}

\maketitle

\begin{abstract}
Face recognition has been extensively studied in computer vision and artificial intelligence communities in recent years. An important issue of face recognition  is data privacy, which receives more and more public concerns.
As a common privacy-preserving technique, Federated Learning is proposed to train a model cooperatively without sharing data between parties. However, as far as we know, it has not been successfully applied in face recognition. This paper proposes a framework named FedFace to innovate federated learning for face recognition. Specifically, FedFace relies on two major innovative algorithms, \textbf{P}artially \textbf{F}ederated \textbf{M}omentum~(PFM) and  \textbf{F}ederated \textbf{V}alidation~(FV).      
PFM locally applies an estimated equivalent global momentum to approximating the centralized momentum-SGD efficiently.
FV repeatedly searches for better federated aggregating weightings via testing the aggregated models on some private validation datasets, which can improve the model's generalization ability.
The ablation study and extensive experiments validate the effectiveness of the FedFace method and show that it is comparable to or even better than the centralized baseline in performance.
\end{abstract}

\section{Introduction}

Face recognition (FR) has wide applications in such as airport check-in and mobile Face ID. With the widespread use of face recognition, concerns of data privacy have been raised. Off-the-shelf FR methods ~\cite{wang2018cosface} ~\cite{deng2019arcface} assume all the data is available within single party and can be trained in a centralized way, while data privacy is not under consideration.



In recent years, federated learning ~\cite{mcmahan2017communication}  has become an important privacy-preserving paradigm in various machine learning tasks\cite{zheng2020federated,zhuang2020performance}, in which model is trained in a distributed way without sharing private data between multi-parties.
However, the potential of federated learning in face recognition is far from being fully exploited. To shatter the concern of data privacy in FR, we propose FedFace to adapt the federated learning framework into the FR problem. Specifically, we consider a cross-silo setting \cite{karimireddy2020mime}, which corresponds to a relatively small number of reliable parties.
Compared with the baseline federated learning framework, \textit{i.e,}, FedAvg, FedFace improves in two aspects.

First, to address the client drift~\cite{zhao2018federated},
in FedFace, we propose the \textbf{P}artially \textbf{F}ederated \textbf{M}omentum~(PFM) algorithm specially designed to fit FR training. Following the idea of MIME~\cite{karimireddy2020mime}, PFM computes statistics globally and applies it locally. Different from MIME computing the gradient for both the local and global models at a local step, PFM estimates the global gradient via the sum of local gradients to keep the training efficiently, which is detailedly described in Sec.~\ref{sec:pfm}. 

Second, the difference among local models is further exploited by evaluating their performance on some validation datasets. In detail, federated learning maintains multi-party models simultaneously, and the difference of local models gives us extra opportunity to make the models more robust via some train-time validations.
However, the privacy of the validation data also needs to be protected.
Therefore, we develop the \textbf{F}ederated \textbf{V}alidation~(FV) algorithm that dynamically searches the near-optimal weightings for federated aggregation via validating the aggregated models on multiple parties, each of which owns a private validation dataset.
FV is detailedly described in Sec.~\ref{sec:fv}.
In summary, our major contributions are:
\begin{itemize}
    \item We propose a novel FR paradigm, federated face recognition, and design the FedFace method to train FR models via federated learning. To our best knowledge,  FedFace is the first to introduce federated learning into the FR community.
    \item We develop Partially Federated Momentum to correct client drift in federated training of FR while keeping the training efficient.
    \item We propose Federated Validation to improve the models' generalization ability in federated learning for FR.
    \item We conduct extensive experiments to evaluate and analyze our method, which validates the effectiveness of the proposed PFM and FV algorithms, and the high performance of FedFace under various experimental settings.
\end{itemize}

\section{Related Work}

\subsection{Face Recognition}
The state-of-the-art training of face recognition consists of two parts: a backbone network that extracts embeddings from training images and a classifier with a softmax-based loss function.  
Recently, some improved softmax-based loss functions,
such as CosFace~\cite{wang2018cosface} and ArcFace~\cite{deng2019arcface}, were proposed to add an angular margin to maximize the inter-class discrepancy and minimize the intra-class variance.
These margin-based methods achieved state-of-the-art performance.
However, existing FR methods do not consider data privacy when training with data from multi-parties.

\subsection{Federated Learning}
Federated Average~(FedAvg)~\cite{mcmahan2017communication} was proposed to learn a shared model by aggregating locally-computed updates.
The convergence of FedAvg was analyzed in many works \cite{stich2018local,pmlr-v119-woodworth20a,khaled2020tighter}. 
\cite{yu2019parallel,DBLP:conf/icml/KoloskovaLBJS20} extended the analysis to heterogeneous clients.
However, the classical FedAvg protects privacy at the cost of performance.

\paragraph{Client drift.}
Client drift was first observed by \cite{zhao2018federated} that local steps lead to ``over-fitting'' to local data when training with non-$i.i.d.$ data.
SCAFFOLD was proposed to correct such drift by using control variates~\cite{pmlr-v119-karimireddy20a}.
MIME~\cite{karimireddy2020mime} generalized SCAFFOLD to all functions, and drew a conclusion that locally applying global momentum is better than the server-only-momentum approaches~\cite{DBLP:conf/iclr/WangTBR20}.
Although MIME performs well, 
    it is harmful to training efficiency,
    because it computes the gradient twice at each local step.
As an improvement, our proposed PFM considers both performance and efficiency simultaneously.

\paragraph{Weighting strategies.}
\cite{8945292} proposed an asynchronous learning strategy on the clients and a temporally weighted aggregation of the local models on the server.
\cite{wu2020fast} assigned different weights for updating the global model based on node contribution adaptively through each training round.
Different from the existing methods, our proposed FV securely searches for good weightings according to the aggregated model's performances on some validation datasets.

\section{Method}\label{sec:meth}

\begin{figure}[!tb]
\centering
\includegraphics[width=0.48\textwidth]{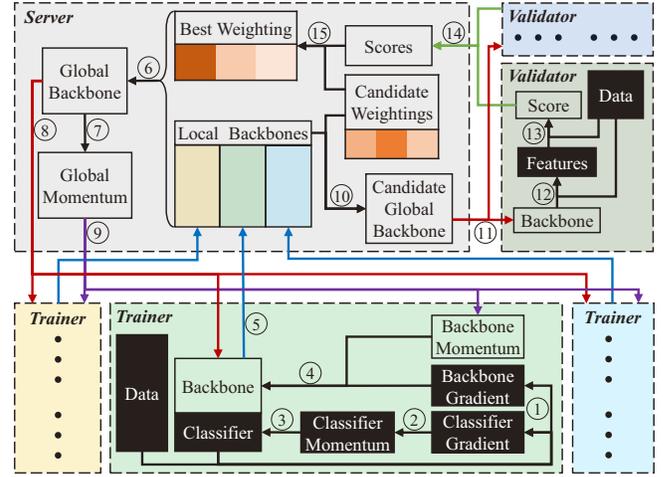}
\caption{
    The framework of FedFace.
    It demonstrates a deployment that consists of a \textit{server}, 3 \textit{trainer}s and 2 \textit{validator}s.
    The blocks with black background are private.
    And the procedures~(arrows) that numbered with 1 to 15 are detailed in Sec.~\ref{sec:meth}
}
\label{fig:fram}
\end{figure}

Fig.~\ref{fig:fram} illustrates the framework of FedFace, which consists of the \textit{server}, the \textit{trainer}s and the \textit{validator}s.
Each \textit{trainer} trains the backbone and the classifier on a set of private training data~(\circled{1} to \circled{4}), and synchronizes its model with the \textit{server} periodically~(\circled{5}).
We denote the set of all \textit{trainer}s as $\mathcal{T}$, and the training data as $X_{1\dots|\mathcal{T}|}$.
The momentum optimizations for the classifier~(\circled{2} and \circled{3}) and  the backbone~(\circled{4}) are different, which are elaborated in Sec.~\ref{sec:pfm}.
The \textit{server} aggregates all the local backbones into a global backbone~(\circled{6}) and maintains the global momentum~(\circled{7}), then sends the global backbone and momentum back to each \textit{trainer}~(\circled{8} and \circled{9}).
Meanwhile, the \textit{server} repeatedly tests random candidate weightings for the aggregation by sending them to the \textit{validator}s~(\circled{10} and \circled{11}), and adjusts the current weighting if a better one is found~(\circled{15}).
Each \textit{validator} scores a model received from the \textit{server} according to its private evaluation data~(\circled{12} and \circled{13}) and sends back the score~(\circled{14}) repeatedly.
We denote the set of all \textit{validator}s as $\mathcal{V}$, and the evaluation data as~$Y_{1\dots|\mathcal{V}|}$.

\subsection{Partially Federated Momentum}\label{sec:pfm}

In the training of face recognition, the feature extractor~(backbone) $f$ with parameter $\Theta$ is shared by all datasets, and the classifiers $c_{1\dots|\mathcal{T}|}$ with parameters $\omega_{1\dots|\mathcal{T}|}$ are different.
Therefore, the momentums for the backbone $M^\Theta$ and classifiers $M^\omega_{1\dots|\mathcal{T}|}$ are maintained separately.

For the classifier on the $i$-th \textit{trainer}, at the $k$-th step of the $r$-th training round, $\omega_{i,r,k}$ is maintained locally with classical momentum\footnote{PyTorch implementation.}:
\begin{equation}
    M^\omega_{i,r,k} = \beta M^\omega_{i,r,k-1} + h_{i,r,k}~\text{,}
\end{equation}
\begin{equation}
    \omega_{i,r,k} = \omega_{i,r,k-1} - \eta_r M^\omega_{i,r,k}
\end{equation}
where $\beta$, $\eta_r$ and $h_{i,r,k}$ are the momentum parameter, the learning rate at the $r$-th round, and the gradient of the classifier on the $i$-th \textit{trainer} at this step, respectively.

And for the backbone, let $\Theta_{r-1}$ and $M^\Theta_{r-1}$ be the last global backbone parameter and global momentum sent to each \textit{trainer} after the $r-1$ training round.
When computing the local backbone parameter $\theta_{i,r,k}$ at the $k$-th step in the $r$-th training round on the $i$-th \textit{trainer}, the global momentum is applying locally:
\begin{equation}
    \theta_{i,r,k} = \theta_{i,r,k-1} - \eta_r\left(g_{i,r,k} + \beta \cdot \frac{M^\Theta_{r-1}}{K}\right)
\end{equation}
where $K$ and $g_{i,r,k}$ are the step number in a training round, and the gradient of the backbone on the $i$-th \textit{trainer} at this step respectively.
After $K$ local steps, the $i$-th \textit{trainer} sends $\theta_{i,r,K}$ to the \textit{server}.
And the \textit{server} aggregates the local backbone parameters to the global parameter $\Theta_r$ with the weighting $w$ that changes over time~(detailed in Sec.~\ref{sec:fv}):
\begin{equation}
   \Theta_r = \sum_{i=1}^{|\mathcal{T}|}w_i \theta_{i,r,K}~\text{.}
\end{equation}
Then, the equivalent global gradient $G_r$ is estimated from the variation between $\Theta_r$ and $\Theta_{r-1}$:
\begin{equation}
   G_r = \frac{\Theta_{r-1} - \Theta_r}{\eta_r} - \beta M^\Theta_{r-1}~\text{.}
\end{equation}
Note that the effects of the learning rate and the locally applied momentum should be eliminated.
Next, the global momentum $M^\Theta_r$ is computed as:
\begin{equation}
   M^\Theta_r = \beta M^\Theta_{r-1} + G_r = \frac{\Theta_{r-1} - \Theta_r}{\eta_r}~\text{.}
\end{equation}
Following \cite{karimireddy2020mime}, the \textit{server} does not apply the global momentum, but sends both the global backbone $\Theta_r$ and the global momentum $M^\Theta_r$ to each \textit{trainer}, and the momentums are applied at the local steps in the next round. 

\begin{figure}[tb]
\centering
\includegraphics[width=0.47\textwidth]{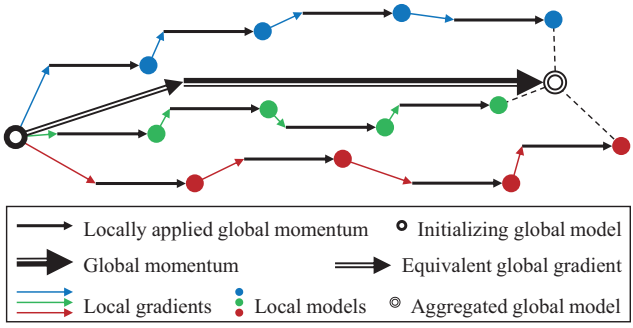}
\caption{The locally applied global momentum.
The global momentum is evenly applied to each step of a training round.
The equivalent global gradient is estimated by subtracting the global momentum from the global model variation.
}
\label{fig:pfm}
\end{figure}

\begin{figure}[tb]
\centering
\begin{tabular}{ccc}
    \includegraphics[width=0.2\textwidth]{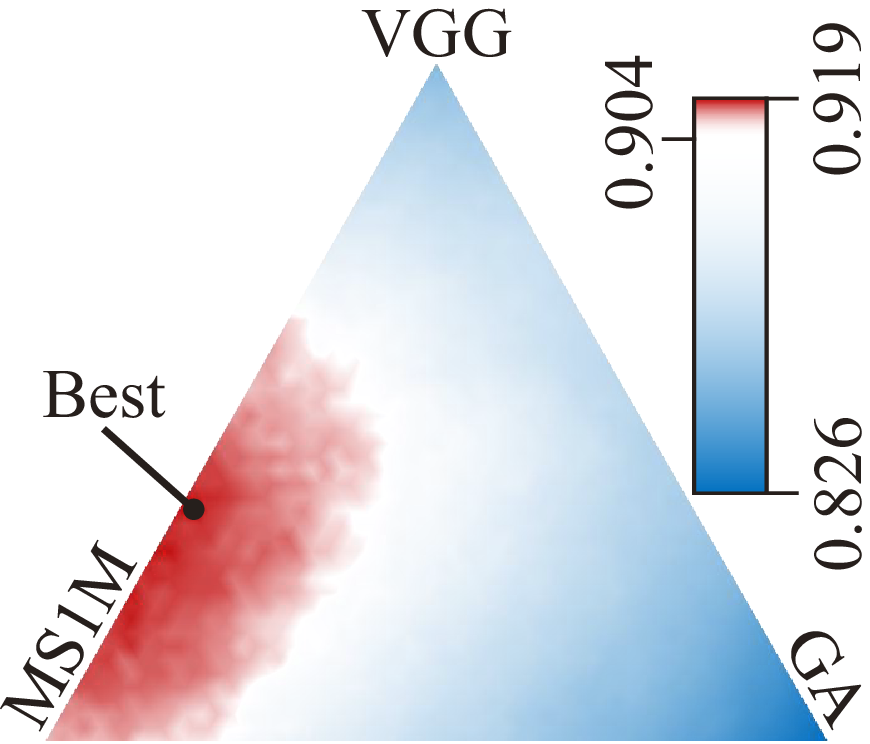} & &
    \includegraphics[width=0.2\textwidth]{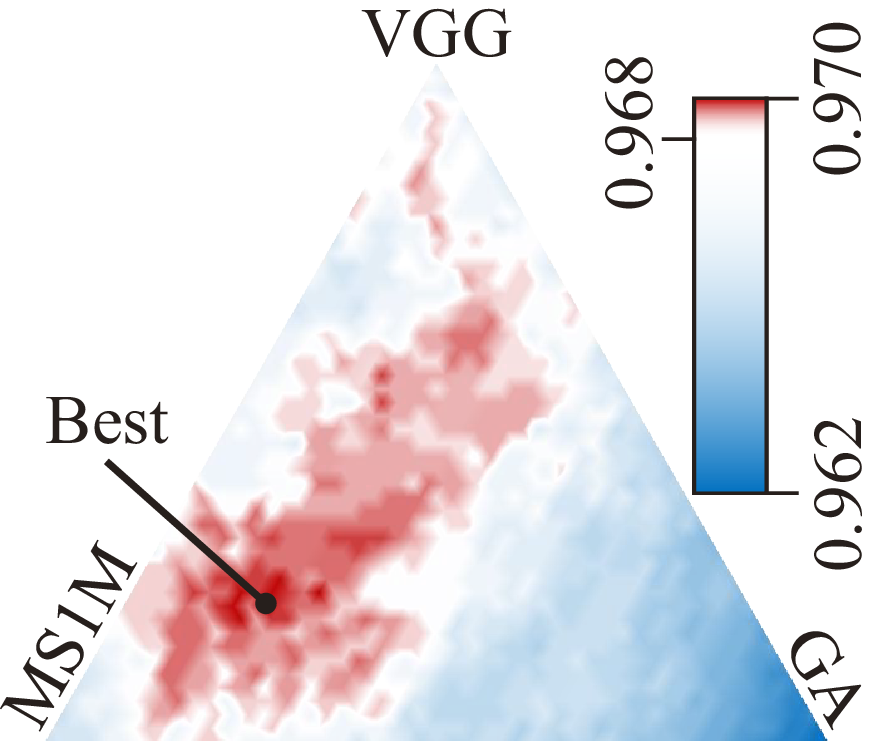} \\
    (a)~Step 12,800 & & (b)~Step 256,000
\end{tabular}
\caption{
    The grid search results on weightings that used for aggregating 3-party models into the global model at different steps.
    The values are the validation scores described in Sec.~\ref{sec:fv} without normalizing.
    The datasets MS1M, GA and VGG are described in Sec.~\ref{sec:data}.
    And the models used for searching are the intermediate checkpoints of the experiment ``FedAvg'' on Fig.~\ref{fig:performance}~(a).
}
\label{fig:fvmot}
\end{figure}

Fig.~\ref{fig:pfm} gives a demonstration of locally applying the global momentum and estimating the equivalent global gradient.
And the complete process of PFM is detailed in Alg.~\ref{alg:pfm}.

\begin{algorithm}[tb]
\caption{Partially Federated Momentum}
\label{alg:pfm}
\textbf{Input}:
    the number of global training rounds $R$,
    the number of local steps per round $K$,
    the backbone model $f$,
    the classifiers $c_{1\dots|\mathcal{T}|}$ of each \textit{trainer},
    the batched training data $X^{1\dots|\mathcal{T}|}_{1\dots R,1\dots K}$ located at each \textit{trainer},
    the initial model $\Theta$,
    the learning rates for each round $\eta_{1\dots R}$,
    the volatile weighting $w$,
    and the momentum parameter~$\beta$
\begin{algorithmic}[1] 
\STATE $M^\Theta \gets \vec{0}$
\FOR{each round $r = 1\dots R$}
    \STATE \textbf{communicate} $\Theta$ to each \textit{trainer}
    \FOR {$i\in 1\dots|\mathcal{T}|$ \textbf{in parallel}}
        \STATE $\theta_i \gets \Theta$
        \FOR{each $k = 1\dots K$}
            \STATE $(g_i, h_i) \gets \nabla c_i\circ f(\theta_i, \omega_i; X^i_{r,k})$
            \STATE $\theta_i \gets \theta_i - \eta_r(g_i + \beta\cdot\frac{M^\Theta}{K})$
            \STATE $M^\omega_i \gets \beta M^\omega_i + h_i$
            \STATE $\omega_i \gets \omega_i - \eta_r M^\omega_i$
        \ENDFOR
        \STATE \textbf{communicate} $\theta^i_t \gets \theta^i_{t,K}$
    \ENDFOR
    \STATE $\Theta' \gets \Theta$
    \STATE $\Theta \leftarrow \sum_{i=1}^{|\mathcal{T}|}w_i \theta_i$
    \STATE $M^\Theta \gets \frac{\Theta' - \Theta}{\eta_r}$
    \STATE \textbf{communicate} $(\Theta, M^\Theta)$ to each \textit{trainer}
\ENDFOR
\end{algorithmic}
\end{algorithm}

\subsection{Federated Validation} \label{sec:fv}

The aggregated global model shows different performance when varying the weightings for aggregation~(see Fig.~\ref{fig:fvmot}).
Intuitively, finding better weightings during training would improve the model performance.



A reasonable idea is to make a full search on some validation data at each synchronization, because training-time validation is helpful in most machine learning tasks to enhance model's generalization ability.
The full search could be a grid search, random search or Bayesian search with lots of steps.
However, it is too slow~(\textit{e.g.}~each grid search on Fig.~\ref{fig:fvmot} spends more than 1 hour with 4 V100 GPUs).
And such kind of accurate search requires extra synchronizations between \textit{trainer}s and \textit{validator}s, which reduces the efficiency of both.
Alternatively, if skipping the synchronization, the search result may be severely obsolete because there are too many training steps between the models used for search and the models applying the search result.
Therefore, we make a trade off here to balance performance and efficiency, that is, asynchronously and repeatedly random searching for $T$ steps.
And $T$ should be a small number because a large $T$ dose not increase the total number of search steps but decreases the timeliness of the search result.


\begin{algorithm}[tb]
\caption{Federated Validation}
\label{alg:fv}
\textbf{Input}:
    validation data $Y_{1\dots|\mathcal{V}|}$ located at each \textit{validator},
    validation functions $\Score_{1\dots|\mathcal{V}|}$,
    the score normalizing function $\Normalize$,
    and volatile local backbones $\theta_{1\dots|\mathcal{T}|}$ stored on the \textit{server}
\begin{algorithmic}[1] 
\WHILE{training is not finished}
    \STATE $\theta' \gets \theta$
    \FOR{$t\in 1\dots T$}
        \STATE $\hat{w}_t \gets$ uniformly sampled weighting \textbf{if} $t>1$ \textbf{else} $w$
        \STATE $\hat\Theta\gets\hat{w}_r^\top\theta'$
        \STATE \textbf{communicate} $\hat\Theta$ to each \textit{validator}
        \FOR {$i\in$ all \textit{validator}s \textbf{in parallel}}
            \STATE \textbf{communicate} $S_{i,t} \gets \Score_i(\hat\Theta;Y_i)$
        \ENDFOR
    \ENDFOR
    \STATE $S \gets \Normalize(S)$
    \STATE $\hat{t} \gets \arg_t\max \sum_{i=1}^{|\mathcal{V}|}S_{i,t}$
    \STATE $w \gets (1-\varphi)w + \varphi\hat{w}_{\hat{t}}$
\ENDWHILE
\end{algorithmic}
\end{algorithm}

In each validation round, we firstly duplicate the current local backbones $\theta$ as $\theta'$ because the parameters should not vary during a whole evaluation round but $\theta$ may be updated when communicating with \textit{trainer}s.
Then, we take $T$ steps of validation.
In the $t$-th step, the candidate weighting $\hat{w}_t$ is generated by randomly sampling.
Exceptionally, $\hat{w}_1$ is the most recent applied weighting, which guarantees the searched result will be at least not worse than the last applied one.
After that, the \textit{server} sends the candidate global parameter 
\begin{equation}
    \hat\Theta=\hat{w}_r^\top\theta'   
\end{equation}
to each \textit{validator} and receives the scores $S$.
The scoring of a validator depends on the corresponding dataset, and it could be an accuracy, a true positive rate, or a loss.
The received scores are then normalized to eliminate the impact of difference in difficulty among these validation datasets.
The normalization could be simply dividing by the standard deviation of the results of an individual \textit{validator} in a single validation round~(named as \textbf{Local Norm}):
\begin{equation}
    S_i' =  \frac{S_i}{\sqrt{\sigma^2(S_i)+\epsilon}}
\end{equation}
or dividing by the statistical moving standard deviation of an individual \textit{validator}~(named as \textbf{Moving Norm}):
\begin{equation}
    \mu = (1-\gamma) \mu_\text{last} + \gamma\overline{S_i}~\text{,}
\end{equation}
\begin{equation}
    \nu = (1-\gamma) \nu_\text{last} + \gamma\cdot\frac{\sum_{s\in S_i}(s-\mu)^2}{|S_i|}~\text{,}
\end{equation}
\begin{equation}
    S_i' =  \frac{S_i}{\sqrt{\nu+\epsilon}}
\end{equation}
where $S_i$, $\gamma$, $\mu$, $\nu$, and $\epsilon$ are the score from the $i$-th \textit{validator}, the norm parameter, the moving mean, the moving variance, and a small number to avoid dividing by zero, respectively.
Finally, the best scored candidate weighting is applied to $w$ with a smooth rate $\varphi$:
\begin{equation}
    \hat{t} = \arg_t\max \sum_{i=1}^{|\mathcal{V}|}S'_{i,t}~\text{,}
\end{equation}
\begin{equation}
    w = (1-\varphi)w_\text{last} + \varphi\hat{w}_{\hat{t}}~\text{.}
\end{equation}

Alg.~\ref{alg:fv} gives a formally description of FV.
And the performances of different normalizing strategies are discussed in~Sec.~\ref{sec:exp}.

\section{Experiments}
\begin{figure*}[tb]
\centering
\small
\begin{tabular}{r@{}ccc}
    & \multicolumn{3}{c}{\includegraphics[height=0.5cm]{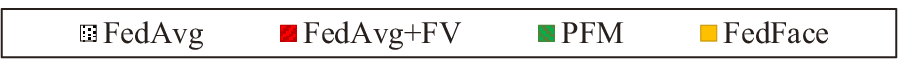}} \\
    \rule{0pt}{2.5ex}
    \belowbaseline[-2mm]{\includegraphics[height=3.75cm, valign=t]{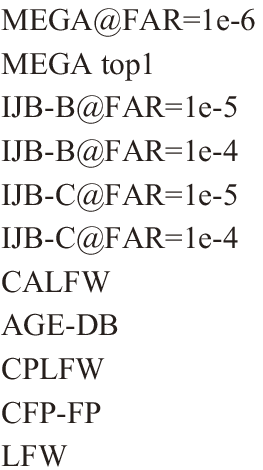}} &
    \includegraphics[height=4.3cm, valign=t]{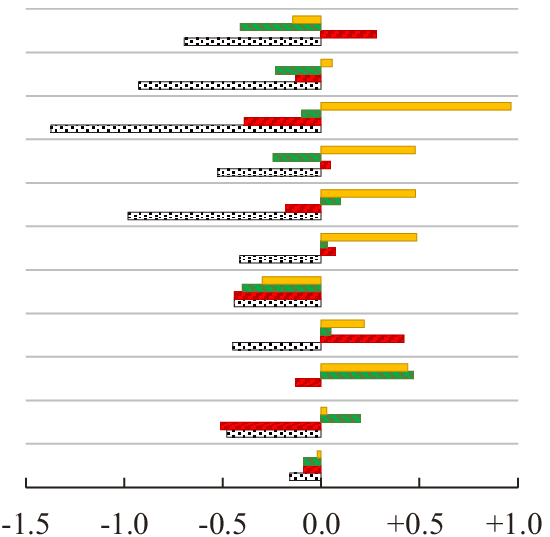} &
    \includegraphics[height=4.3cm, valign=t]{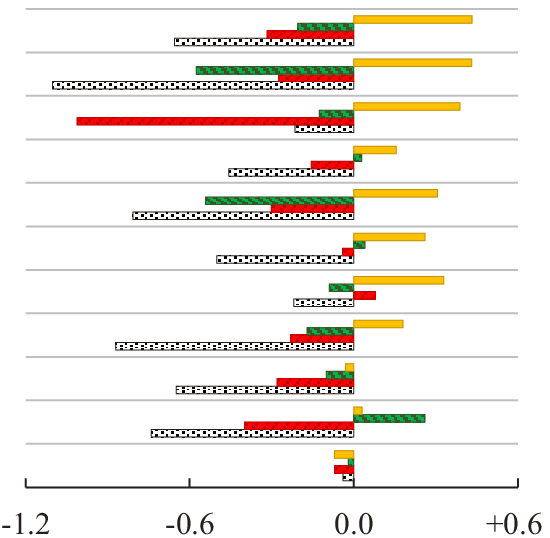} &
    \includegraphics[height=4.3cm, valign=t]{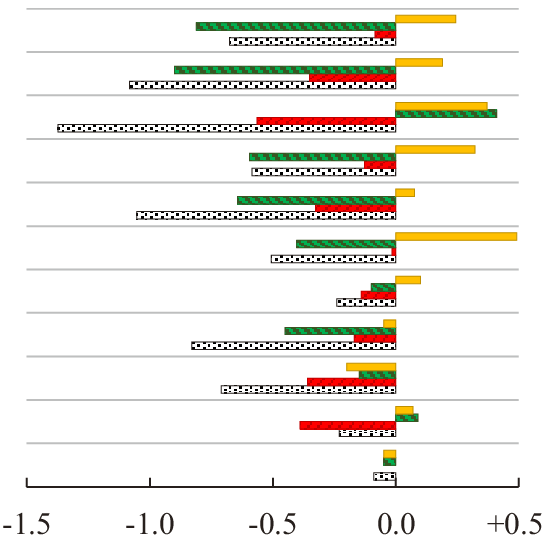} \\
    \rule{0pt}{2.5ex} & 
    (a)~IR-18-AF, 3-party, $K=100$ & 
    (b)~IR-18-AF, 3-party, $K=400$ &
    (c)~IR-18-AF, 3-party, $K=1600$ \\
    \rule{0pt}{5ex}
    \belowbaseline[-2mm]{\includegraphics[height=3.75cm, valign=t]{figures/scry4.eps}} &
    \includegraphics[height=4.3cm, valign=t]{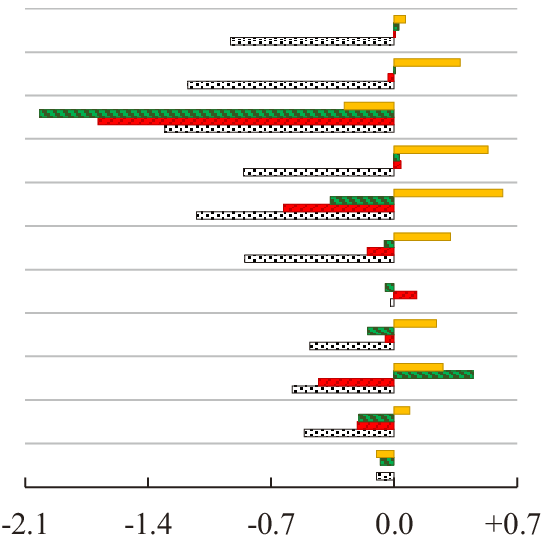} &
    \includegraphics[height=4.3cm, valign=t]{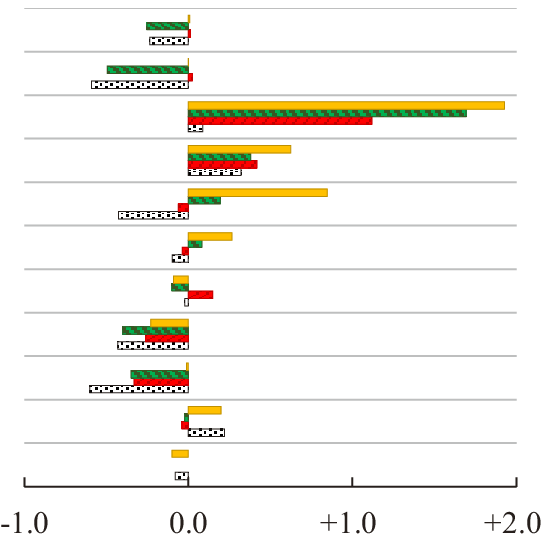} &
    \includegraphics[height=4.3cm, valign=t]{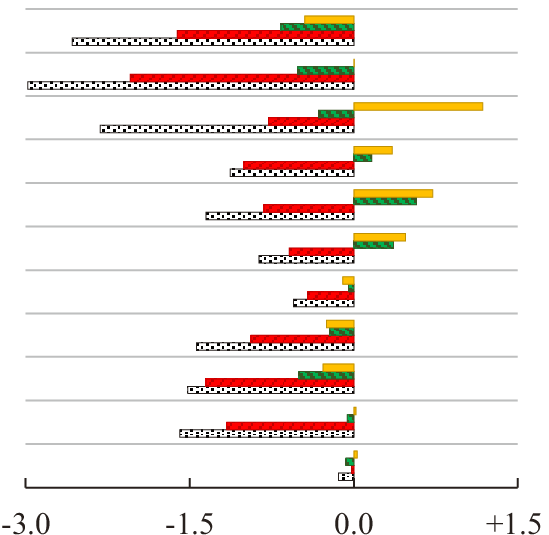} \\
    \rule{0pt}{2.5ex} & 
    (d)~IR-18-CF, 3-party, $K=100$ & 
    (e)~IR-34-AF, 3-party, $K=100$ &
    (f)~IR-18-AF, 12-party, $K=100$ 
\end{tabular}
  \caption{
    Relative performance comparison with the centralized baseline. 
    The caption of each sub-figure describes the corresponding model name, party size and synchronization interval~($K$).
    The suffixes ``-AF'' and ``-CF'' in the model names denote for loss functions ArcFace and CosFace respectively.
    Each value is the result of a certain model subtracted by that of the corresponding centralized baseline, in percentage.
  }
\label{fig:performance}
\end{figure*}

\subsection{Datasets}\label{sec:data}
\paragraph{Training Sets.}
We employ refined MS1MV2~(MS1M in short), Asian-DeepGlint~\cite{deng2019arcface}~(GA in short), and VGGFace-2~\cite{cao2018vggface2}~(VGG in short) as training data. MS1MV2 is refined from MS-Celeb-1M dataset~\cite{guo2016ms} and contains about 5.8M images of 85K individuals.  Asian-DeepGlint is a representative Asian face dataset which includes 2.8M images and 94K identities. 
VGGFace-2 
has 9K subjects while each subject owns an average of 362.6 images.
These datasets are stored at different \textit{trainer}s in our experiments.

\paragraph{Validation Sets.}
We apply Federated Validation on several popular benchmarks, including LFW~\cite{lfw},  CPLFW~\cite{CPLFWTech}, CALFW~\cite{zheng2017crossage}, CFP-FP~\cite{cfp-fp}, AgeDB~\cite{moschoglou2017agedb}. LFW is the most common face verification test dataset, which contains 13,233 web-collected images from 5,749 different identities. CPLFW, CALFW, CFP-FP, and AgeDB focus on the performance of large variations in pose and age.

\paragraph{Evaluation Sets.}
MegaFace~\cite{kemelmacher2016megaface} (MEGA in short) is the most representive challenging open testing protocol.
The gallery set of MegaFace includes 1M images of 690K subjects, and the probe set from FaceSurb includes 100K photos of 530 unique individuals. 
The IJB-B~\cite{whitelam2017iarpa} and IJB-C ~\cite{maze2018iarpa} are introduced as two large-scale face verification protocols. IJB-B provides 12,115 templates with 10,270 genuine matches and 8M impostor matches. IJB-C further provides 23,124 templates with 19,557 genuine and 15,639K impostor matching pairs.

\subsection{Implementation Details}\label{sec:impl}
We implement the proposed method based on PyTorch1.6.

\paragraph{Backbones.}
We use the adapted ResNet18 with improved residual unit (IR-18)~\cite{deng2019arcface} as the backbone network, which has better convergence in early training stage. We also adopt a bigger backbone IR-34 for ablation study.
\paragraph{Loss functions.}
We implement ArcFace and CosFace using model parallel for massive identities training. In ArcFace loss, we follow the common setting to set the feature scale $s = 64$ and the margin parameter $m = 0.5$. And in CosFace loss, we set $s = 64$ and $m = 0.35$.

\paragraph{Training.}
We train the models on Nvidia V100 GPUs.
For the 3-party experiments, MS1M, GA and VGG are trained on 8, 4, and 4 GPUs respectively.
All training face images are cropped to $112 \times 112 $ according to five facial points. 
For the 12-party experiments, each of the 3 training datasets is evenly split into 4 parts,
and each part of MS1M, GA and VGG are trained on 2, 1, and 1 GPU respectively.
The batch size of each GPU is set to 64.
We train all the models up to 26 epochs, and the number of steps per epoch is set to 12,800.
The learning rate is set to 0.1 at initialization, and decays by 0.1 at the end of epoch 10, 18 and 24.
The momentum parameter $\beta$ is set to 0.9.

\paragraph{Validation.}
Each of the above described validation dataset contains a public 10-fold split.
We use the first 5 folds for federated validation and the others for performance evaluation, and vice versa.
When validating, for each fold in the 5 folds, the \textit{validator} selects the threshold by verifying on the remaining 4 folds to calculate accuracy.
And the evaluation score of each dataset is the average of the corresponding 5 folds.
We set the hyper-parameters $\epsilon=0.001$, $\varphi=0.01$, $\gamma=0.01$ and $T=3$.

\paragraph{Performance evaluation.}
Following the mainstream face recognition works, we evaluate the True Acceptance Rate (TAR) on IJB-B and IJB-C at False Acceptance Rate (FAR) equals to 1e-4 and 1e-5 respectively.
For MegaFace, we evaluate the top 1 accuracy for identification, and TAR at FAR~=~1e-6 after data refinement on both prob set and distractors.
For the 5 datasets that used for validation, we evaluate the remaining halves.
And for each one of the evaluating folds, the other 9 folds are used to select the threshold.
All the experiments that use FV are done twice for fair comparison, and the average results are reported.

\subsection{Performance Comparison }\label{sec:exp}

As shown in Fig.~\ref{fig:performance}, we conduct performance comparison under 6 different settings. 
For each setting, we train 5 models: centralized training (baseline), FedAvg, FedAvg with FV, PFM, and PFM with FV (\textit{a.k.a.} FedFace). 
And relative result is reported by subtracting the result of centralized baseline from the result of each of the 4 other methods.

\paragraph{Results under different synchronization intervals.}
The smaller the synchronization interval, the smaller the impact of client drift.
However, the synchronization interval cannot be set too small due to the limitation of bandwidth.
Therefore, we evaluate the methods under different synchronization intervals.
In Fig.~\ref{fig:performance}~(a), (b) and (c), 
the synchronization interval $K$ is set to 100, 400 and 1600 respectively.
The results show that under the 3 settings:
(1) FedAvg suffers  significant performance degradation, and the degradation becomes worse with the increase of $K$;
(2) FedAvg+FV performs a little better than pure FedAvg;
(3) Comparing to FedAvg, PFM performs more closely to the centralized baseline;
(4) FedFace performs better than the baseline in most benchmarks for all the 3 settings, which means that PFM and FV are complementary to each other.

\paragraph{Results under different backbones and loss functions.}
To verify the universality of the proposed method, we also conduct experiments under different backbones and loss functions.
In Fig.~\ref{fig:performance}~(d), the loss function is replaced by CosFace.
And in Fig.~\ref{fig:performance}~(e), the backbone is replaced by IR-34.
PFM is comparable to the baseline, and FedFace performs better in both,
which again supports the effectiveness and complementarity of PFM and FV.

\paragraph{Results under different party numbers.}
As the party number grows, the drift problem get worse.
Therefore, we compare the methods under a 12-party setting.
The results are shown in Fig.~\ref{fig:performance}~(f).
FedAvg gets worse performance, but PFM is still comparable to the baseline.
And FedFace performs a little better than the baseline.

\subsection{Ablation Study}
\paragraph{The effect of PFM on training loss.}

\begin{table}[tb]
\centering
\small
\begin{tabular}{lccccc}
\toprule
Model & Party & Solo & Centralized & FedAvg & PFM \\
\midrule
\multirow{3}{*}{\makecell{ IR-18\\ArcFace }} 
& MS1M & 4.21 & 4.63 & 5.44 & 4.89 \\
& GA & 5.78 & 4.61 & 5.17 & 3.88 \\
& VGG & 6.11 & 5.70 & 6.15 & 5.25 \\
\midrule
\multirow{3}{*}{\makecell{ IR-18\\CosFace }} 
& MS1M & 1.71 & 2.06 & 2.29 & 1.97 \\
& GA & 1.86 & 1.97 & 2.25 & 1.63 \\
& VGG & 2.89 & 2.73 & 2.96 & 2.56 \\
\midrule
\multirow{3}{*}{\makecell{ IR-34\\ArcFace }} 
& MS1M & 2.27 & 3.03 & 3.79 & 2.96 \\
& GA & 2.61 & 1.71 & 2.11 & 2.18 \\
& VGG & 1.84 & 1.57 & 1.48 & 1.21 \\
\bottomrule
\end{tabular}
\caption{
    Final training loss values of different models on the 3 datasets.
    ``Solo'' means the model is trained with only one dataset.
    The values are exponential smoothed with a factor of 0.99 for stability.
}
\label{tab:loss}
\end{table}
\begin{figure}[tb]
\centering
\begin{tabular}{r@{}l}
    \belowbaseline[-2mm]{\includegraphics[height=4.9cm, valign=t]{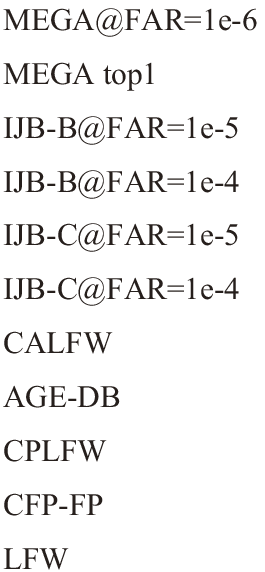}} &
    \includegraphics[height=5.5cm, valign=t]{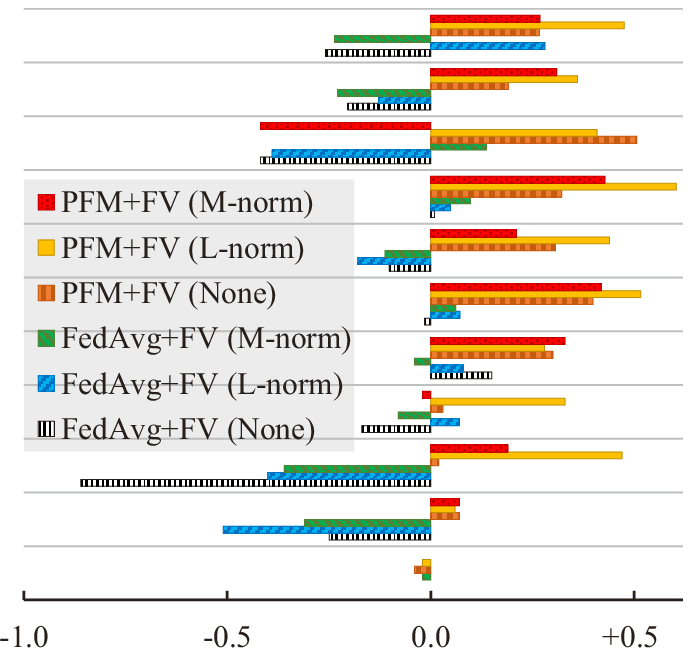}
\end{tabular}
\caption{
    The performance of different score normalizing strategies in FV.
    ``M-norm'' denotes Moving Norm, 
``L-norm'' denotes Local Norm, 
    and ``None'' means not to normalize but to use the original scores.
}
\label{fig:norm}
\end{figure}

As shown in Tab.~\ref{tab:loss}, we check the final training loss values of different models on the 3 datasets under the setting of 3-party and $K=100$.
We can see: 
(1)~The loss of centralized training is larger than solo training on MS1M, but smaller on GA and VGG, this is because MS1M is a dataset relatively easier to learn  than the other two;
(2)~The loss of FedAvg is larger than centralized training due to client drift;
(3)~The loss of PFM is significantly smaller than FedAvg, and even smaller than centralized training in most cases, which shows that PFM can considerably enhance models' fitting ability in federated learning.


\paragraph{Performance of different norm strategies for FV.}
We evaluate two above mentioned score normalization strategies for FedAvg+FV and PFM+FV.
Experiments are performed under IR-18 backbone, 3-party, and $K=100$.
The results in Fig.~\ref{fig:norm} show that Local Norm always performs best.
Therefore, Local Norm is used in all the other experiments.

\paragraph{The variations of weightings in FV.}
\begin{figure}[tb]
\centering
\includegraphics[width=0.48\textwidth]{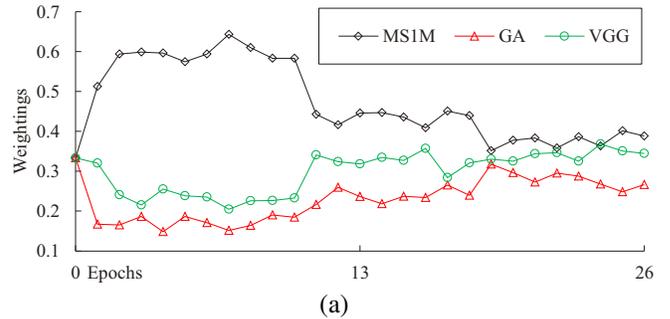} \\
(a) \\
\vspace{2mm}
\includegraphics[width=0.48\textwidth]{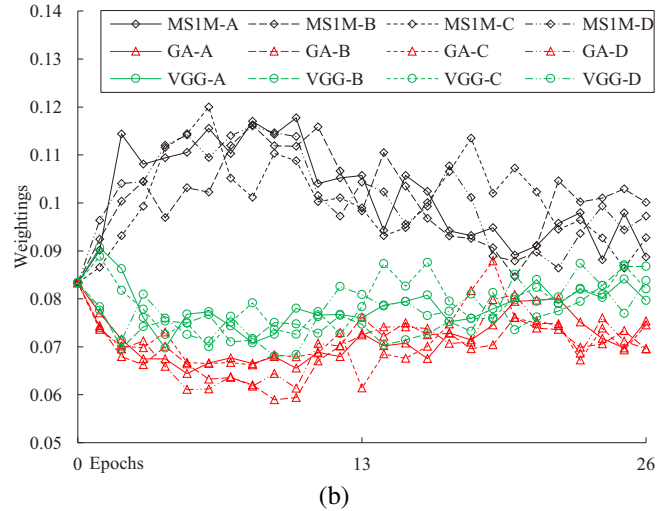} \\
(b)
\caption{The variations of weightings $w$ for each party when performing Federated Validation.
(a) The weightings for a 3-party experiment.
(b) The weightings for a 12-party experiment.
The quarter split parts of each dataset are labeled with a `-A', `-B', `-C' and `-D' suffix respectively.
}
\label{fig:weight}
\end{figure}
As shown in Fig.~\ref{fig:weight}, the weightings applied in FV show similar pattern in both the 3-party and the 12-party experiments.
The parties that use MS1M data are assigned with larger weightings in the early stages of the training.
And the weightings tend to average as the model converges.

\section{Conclusion}
In this work, we propose federated face recognition  to train face recognition models using multi-party data via federated leaning to avoid privacy risks.
We develop the PFM and FV algorithms to improve the performance of federated face recognition. 
Extensive experiments are conducted to evaluate the proposed method. Experimental results show that the models trained with our method are comparable to or even better than the centralized baseline under various hyper-parameter settings.
In the future, we plan to adapt the proposed method to serve more federated learning tasks.

\bibliographystyle{named}
\bibliography{FV}

\end{document}